%% file: main.tex
\documentclass[10pt,twocolumn,letterpaper]{article}

\usepackage{caption}
\usepackage{titling}
\usepackage{iccv}
\usepackage{times}
\usepackage{epsfig}
\usepackage{graphicx}
\usepackage{amsmath}
\usepackage{amssymb}
\usepackage{mathtools}
\usepackage{booktabs}
\usepackage{algorithm}
\usepackage{algpseudocode}
\usepackage{balance}
\usepackage{multirow}

\usepackage[accsupp]{axessibility} 

\usepackage{algorithmicx}
\usepackage[breaklinks=true,bookmarks=false]{hyperref} 

\iccvfinalcopy

\def\iccvPaperID{2605}

\ificcvfinal\fi

\DeclarePairedDelimiter\abs{\lvert\lvert}{\rvert\rvert}

\newcommand{\MYhref}[3][blue]{\href{#2}{\color{#1}{#3}}}

\newcommand{\ba}{\mathbf{a}}
\newcommand{\bb}{\mathbf{b}}

\newcommand{\Etal}   {\textit{et al.}}

\begin{document}

\title{Self-Calibrating Neural Radiance Fields}

\author{
Yoonwoo Jeong$^1$~~~~~~Seokjun Ahn$^1$~~~~~~Christopher Choy$^2$\\
Animashree Anandkumar$^{2,3}$~~~~~~Minsu Cho$^1$~~~~~~Jaesik Park$^1$ 
\vspace{3mm}\\
\mbox{POSTECH$^1$~~~~~~NVIDIA$^2$~~~~~~Caltech$^3$ }
}

 \twocolumn
 [{
 \renewcommand\twocolumn[1][]{#1}
 \maketitle
 \begin{center}
 \centering
 \vskip0.1cm
 \centering
 \vspace{-4mm}
 \begin{minipage}{0.27\linewidth}
 \includegraphics[width=\textwidth]{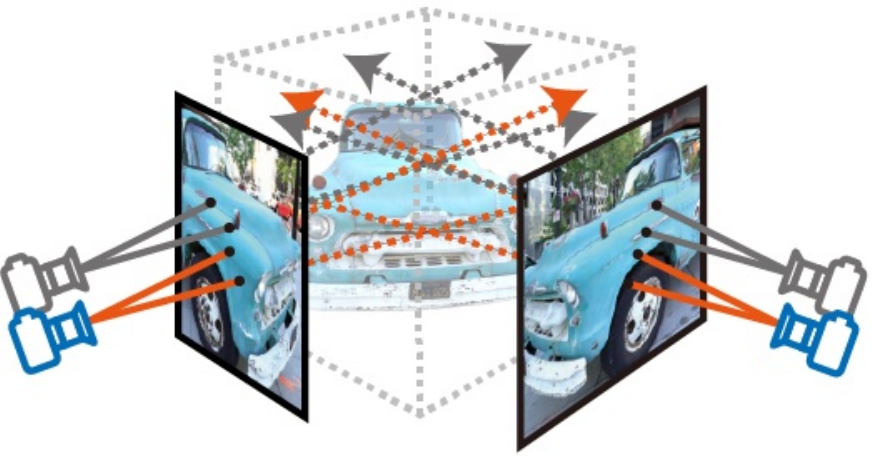}
 \centering\small{(a) Calibrating extrinsic\\camera parameters}\vspace{1mm}
 \end{minipage}\hfill
 \begin{minipage}{0.27\linewidth}
 \includegraphics[width=\textwidth]{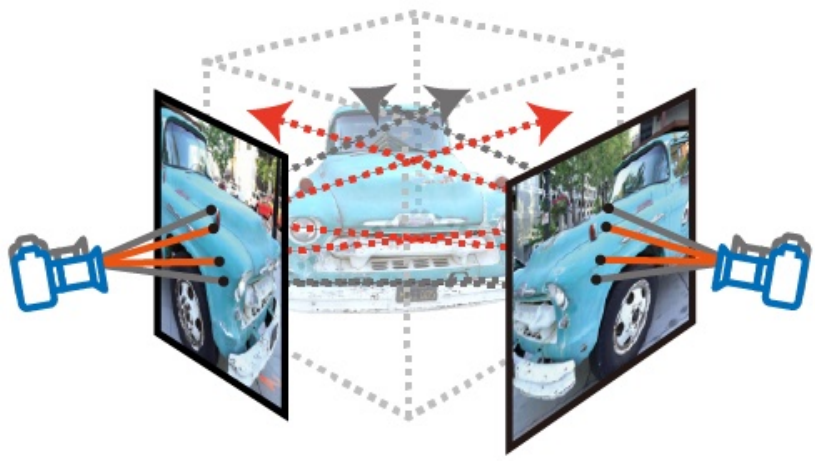}
 \centering\small{(b) Calibrating intrinsic\\camera parameters}\vspace{1mm}
 \end{minipage}\hfill
 \begin{minipage}{0.27\linewidth}
 \includegraphics[width=\textwidth]{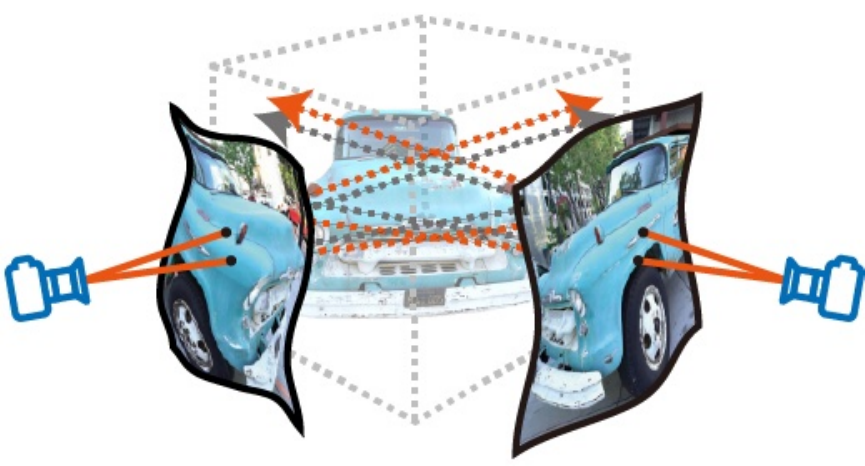}
 \centering\small{(c) Calibrating non-linear\\distortion parameters}
 \end{minipage}\hfill
  \begin{minipage}{0.15\linewidth}
 \includegraphics[width=\textwidth]{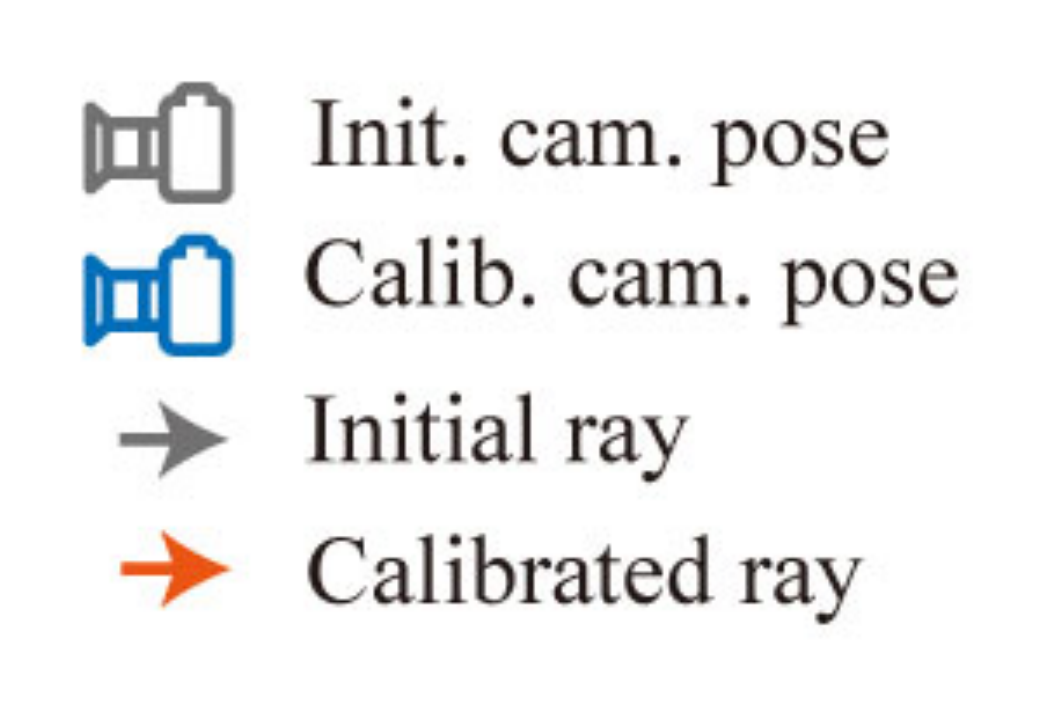}
 \centering
 \end{minipage}\hfill
 \vspace*{3mm}
 \captionof{figure}{
 We visualize calibration processes of three different  camera parameters using our method. Our method calibrates extrinsic camera parameter (a), intrinsic camera parameters (b), non-linear camera model noise (c). \vspace{3mm}
 }
 \label{fig:teaser}
 \end{center}%
 }
]

\input{sections/0_abstract}
\input{sections/1_intro}
\input{sections/2_related}
\input{sections/3_preliminary}
\input{sections/4_camera_model}
\input{sections/5_geometric_and_photometric}
\input{sections/6_optimization}

\input{sections/7_experiment}
\input{sections/conclusion.tex}

\clearpage

\newpage
\balance
{\small
\bibliographystyle{ieee_fullname}
\bibliography{egbib}
}

\pagebreak
\setcounter{section}{0}
\setcounter{table}{0}
\setcounter{figure}{0}

\title{Self-Calibrating Neural Radiance Fields: Supplementary Materials}
\author{}
\maketitle

\input{supplementary/implementation_details}
\input{supplementary/llff_with_colmap}
\input{supplementary/llff_wo_colmap}
\input{supplementary/npp_qual}
\input{supplementary/npp_fisheye}
\input{supplementary/llff_ablation}

\input{supplementary/qualitative}

\end{document}

%% file: sections/0_abstract.tex
\vspace{2mm}
\begin{abstract}
\vspace{-4mm}

In this work, we propose a camera self-calibration algorithm for generic cameras with arbitrary non-linear distortions. We jointly learn the geometry of the scene and the accurate camera parameters without any calibration objects. Our camera model consists of a pinhole model, a fourth order radial distortion, and a generic noise model that can learn arbitrary non-linear camera distortions. While traditional self-calibration algorithms mostly rely on geometric constraints, we additionally incorporate photometric consistency. This requires learning the geometry of the scene, and we use Neural Radiance Fields (NeRF). We also propose a new geometric loss function, viz., projected ray distance loss, to incorporate geometric consistency for complex non-linear camera models. We validate our approach on standard real image datasets and demonstrate that our model can learn the camera intrinsics and extrinsics (pose) from scratch without COLMAP initialization. Also, we show that learning accurate camera models in a differentiable manner allows us to improve PSNR over baselines. 
Our module is an easy-to-use plugin that can be applied to NeRF variants to improve performance. The code and data are currently available at \MYhref[magenta]{https://github.com/POSTECH-CVLab/SCNeRF}{https://github.com/POSTECH-CVLab/SCNeRF}

\end{abstract}
\vspace{-6mm}

%% file: sections/1_intro.tex
\section{Introduction}

Camera calibration is one of the crucial steps in computer vision. Through this process, we learn how the incoming rays map to pixels and thus connect the images to the physical world. Thus, it is a fundamental step in many applications such as autonomous driving, robotics, augmented reality, and many more.

Camera calibration is typically done by placing calibration objects (e.g., a checkerboard pattern) in the scene and estimating the camera parameters using the known geometry of the calibration objects.
However, in many cases, calibration objects are not readily available and can interfere with the perception tasks when cameras are deployed in the wild. 
Thus, calibrating without any external objects, or self-calibration, has been an important research topic; first proposed in Faugeras~\Etal~\cite{faugeras1992camera}. The paper has spurred many follow-ups, some of which propose to globally optimize or embed constraints into the self-calibration optimization process~\cite{pollefeys1999stratified,zeller1996camera,chandraker2007autocalibration,chandraker2010globally}.

Although there has been much progress in developing self-calibration algorithms, all these methods have limitations:
1) the camera model used in self-calibration is a simple linear pinhole camera model. This camera-model design cannot incorporate generic non-linear camera noise that is prevalent in all commodity cameras resulting in less accurate camera calibration. 2) self-calibration algorithms use only a sparse set of image correspondences, and direct photometric consistency has not been used for self-calibration. 3) they use correspondences from a non-differentiable process and do not improve the 3D geometry of the objects, which could improve the camera model. Let us discuss each limitation in detail.

First, a linear pinhole camera model can be formulated as $\mathbf{K}\mathbf{x}$ where $\mathbf{K}\in \mathbb{R}^{3 \times 3}$ and $\mathbf{x}$ is a homogeneous 3D coordinate. This linear model can simplify a camera model and computation, but real lenses have complex non-linear distortions which allow capturing accurate mapping between the real world and images~\cite{grossberg2001general,ramalingam2016unifying,sturm2003generic,schops2020having}. However, traditional self-calibration algorithms assume linear camera models for computational efficiency at the cost of accuracy.

Second, conventional self-calibration methods solely rely on the geometric loss or constraints based on the epipolar geometry, such as Kruppa's method~\cite{kruppa1913ermittlung,hartley1997kruppa, kruppa} that only uses a set of sparse correspondences extracted from a non-differentiable process. This could lead to diverging results with extreme sensitivity to noise when a scene does not have enough interest points. On the other hand, photometric consistency is a physically-based constraint that forces the same 3D point to have the same color in all valid viewpoints. It can create a large number of physically-based constraints to learn accurate camera parameters.

Lastly, conventional self-calibration methods use an off-the-shelf non-differentiable feature matching algorithm and do not improve or learn the geometry. It is well known that the better we know the geometry of the scene, the more accurate the camera model gets. This fact is essential since the geometry of the scene is the sole source of input for self-calibration.

In this work, we propose a self-calibration algorithm for generic camera models that end-to-end learn parameters for the basic pinhole model and radial distortion and non-linear camera noise. For this, our algorithm jointly learns geometry together with a unified end-to-end differentiable framework that allows better geometry to improve camera parameters.
In particular, we use the implicit volumetric representation or Neural Radiance Fields~\cite{mildenhall2020nerf} for the differentiable scene geometry representation.

We also propose a geometric consistency designed for our camera model and train the system together with the photometric consistency for self-calibration, which provides a large set of constraints.
The novel geometric consistency forces rays from corresponding points on images to be close to each other, which overcomes the pinhole camera assumption in the conventional geometric losses derived from Kruppa's method~\cite{kruppa1913ermittlung,hartley1997kruppa, kruppa} for self-calibration~\cite{zeller1996camera}. 

Experimentally, we show that our models can learn camera parameters, including intrinsics and extrinsics, without the standard COLMAP initialization. Also, when the initialization values for these camera parameters are given, we fine-tune the camera parameters accurately, which improves the underlying geometry and novel view synthesis.
We test our model on fish-eye images with COLMAP learned camera radial distortion parameters to analyze the distortion model and show that our model outperforms the baselines by a significant margin.
In addition, we show that our non-linear camera model is modular and can be applied to NeRF variants such as NeRF~\cite{yu2020pixelnerf} and NeRF++~\cite{zhang2020nerf++}.

%% file: sections/2_related.tex
\section{Related Work}

\noindent\textbf{Camera Distortion Model.}
Traditional 3D vision tasks often assume that the camera model is a simple pinhole model. With the development of camera models, various camera models have been introduced, including fish-eye models, per-pixel generic models. Although per-pixel generic models are more expressive, they are difficult to optimize. Schops \Etal~\cite{Schops_2020_CVPR} propose a model locating between 12 parameters and per-pixel generic models. They have shown the proposed model has less reprojection error than other camera models. Strum and Srikumar~\cite{sturm2003generic} propose several methods that show how to calibrate a general imaging model, where structures are known, but viewpoints are unknown. The proposed methods allow learning central cameras without using any distortion model. Grossberg and Nayar ~\cite{grossberg2001general} propose a general imaging model that uses virtual sensing elements that describes the mapping between incoming ray and pixel. They also propose a calibration method that finds parameters of virtual sensing elements and shows that the method can be applied to any imaging system. Ramalingam and Sturm~\cite{ramalingam2016unifying} interpret the camera model as a function that maps pixel to a 3D ray. With this interpretation, they model various cameras, such as central cameras or axial cameras. 

\vspace{2mm}
\noindent\textbf{Camera auto-calibration} is the process of estimating camera parameters from a set of uncalibrated images and cameras without using external calibration objects in the scene, such as checkerboard patterns. Zeller~\Etal~\cite{zeller1996camera} propose a self-calibration method that adopts the Kruppa equation to self-calibrate the camera parameters in a video sequence. Pollefeys~\Etal~\cite{pollefeys1999stratified} propose a stratified method for calibration using modulus constraints. Chandraker~\Etal~\cite{chandraker2007autocalibration} propose a self-calibration algorithm that incorporates the rank and positive semi-definite constraints into the optimization. Chandraker~\Etal~\cite{chandraker2010globally} incorporate the branch and bound method for the globally optimal stratified self-calibration algorithm. Ha~\Etal~\cite{ha2016high} adopt a loss, which implicitly calibrates the camera models using the correspondences between image pairs to produce a high-quality depth map from uncalibrated small motion clips. Engel ~\Etal~\cite{engel2016photometrically} propose a novel approach to calibrate the response function and the non-parametric vignetting function to generate a more accurate tracking model.
\vspace{2mm}
\noindent\textbf {Novel View Synthesis.}
Neural Radiance Fields~\cite{mildenhall2020nerf} synthesize novel views by learning volumetric scene function with multi-layer perceptron. Several improvements on Neural Radiance Field have been proposed. Zhang~\Etal ~\cite{zhang2020nerf++} improve the original NeRF model by discriminating background and foreground. Liu~\Etal ~\cite{liu2020neural} propose a sparse voxel field approach that skips ray marching of the voxels containing no relevant contents, enabling efficient and more precise rendering. Yariv~\Etal~\cite{yariv2020universal} synthesize novel views by reconstructing the 3D surface as a level set of signed distance functions with a neural network. However, surface-based rendering requires a binary mask distinguishing background and foreground.
Moreover, it is not suitable to reconstruct real scenes since the model also reconstructs the background surface. Yu~\Etal~\cite{yu2020pixelnerf} propose a learning framework to learn scene information using few images. Yen~\Etal~\cite{yen2020inerf} address an inverse problem of NeRF, which estimates poses of observed images. They've used test images to predict the poses of the test images and re-trained the NeRF network with the predicted poses for better rendering quality. 

%% file: sections/3_preliminary.tex
\section{Preliminary}

We use the neural radiance fields to learn the 3D scene geometry, which is crucial for learning the photometric loss for self-calibration. In this section, we briefly cover the definitions of the neural radiance fields: NeRF~\cite{mildenhall2020nerf} and NeRF++~\cite{liu2020neural}.

\vspace{2mm}
\noindent\textbf{Implicit Volumetric Representation.} Learning dense 3D geometry of a scene using an implicit representation recently gained significant attention due to its robustness and accuracy. It learns two implicit representations: transparency $\alpha(\mathbf{x})$ and color $\mathbf{c}(\mathbf{x}, \mathbf{v})$ where $\mathbf{x} \in \mathbb{R}^3$ is the 3D position in the world coordinate, $\mathbf{r}_d \in \{\mathbf{r}_d | \mathbf{r}_d \in \mathbb{R}^3, |\mathbf{r}_d| = 1\}$ is a normal 3-vector representing the direction of a ray $\mathbf{r}(t) = \mathbf{r}_o + t \mathbf{r}_d$.

The color value $\mathbf{C}$ of a ray can be represented as an integral of all colors weighted by the opaqueness along a ray, or can be approximated as the weighted sum of colors at $N$ points along a ray.
\begin{equation}
    \hat{\mathbf{C}}(\mathbf{r}) \approx \sum_i^N \left( \prod_{j = 1}^{i - 1} \alpha(\mathbf{r}(t_j), \Delta_j) \right) (1 - \alpha(t_i, \Delta_i)) \mathbf{c}(\mathbf{r}(t_i), \mathbf{v})\label{eq:nerf_volumetric_rendering}
\end{equation}
where $\Delta_i = t_{i+1} - t_i$. Thus, the accuracy of the method depends highly on the number of samples as well as how we sample points.

\vspace{2mm}
\noindent\textbf{Background Representation with Inverse Depth.}
The volumetric rendering used in NeRF is effective and robust if the space the network to capture is bounded. However, on the outdoor scene, the volume of the space is unbounded, and the number of samples required to capture the space increases proportionally, often computationally prohibitive. Instead, Zhang~\Etal~\cite{zhang2020nerf++} propose NeRF++ to model foreground and background with separate implicit networks while the background ray is reparametrized to have bounded volume. The network architecture of ~\cite{zhang2020nerf++} can be succinctly formulated as two implicit networks: one for foreground and one for background. In this paper, we will explore both NeRF and NeRF++ to analyze our camera self-calibration model.

%% file: sections/4_camera_model.tex
\section{Differentiable Self-Calibrating Cameras}
\label{sec:camera_model}

In this section, we provide the definition of our differentiable camera model that combines the pinhole camera model, radial distortion, and a generic non-linear camera distortion for self-calibration~\cite{schops2020having}. 
Mathematically, a camera model is a mapping $\mathbf{p}=\pi(\mathbf{r})$ that defines a 3D ray $\mathbf{r}$ to a 2D coordinate $\mathbf{p}$ in the image plane. In this work, we focus on the unprojection function, or a ray, $\mathbf{r}(\mathbf{p}) = \pi^{-1}(\mathbf{p})$ as the geometry learning and our projected ray distance only requires the unprojection of a pixel to a ray.
Thus, we use the term camera model and camera unprojection interchangeably, and we represent a ray $\mathbf{r}(\mathbf{p})$ of a pixel $\mathbf{p}$ as a pair of 3-vectors: a direction vector $\mathbf{r}_d$ and an offset or a ray origin vector $\mathbf{r}_o$.

Our camera unprojection process consists of two components: unprojection of pixels using a differentiable pinhole camera model and generic non-linear ray distortions. We first mathematically define each component.

\subsection{Differentiable Pinhole Camera Rays}

The first component of our differentiable camera unprojection is based on the pinhole camera model, which maps a 4-vector homogeneous coordinate in 3D space to a 3-vector in the image plane.

First, we decompose the camera intrinsics into the initialization $K_0$ and the residual parameter matrix $\Delta K$. This is due to the highly non-convex nature of the intrinsics matrix that has a lot of local minima. Thus, the final intrinsics is the sum of these $K = K_0 + \Delta K \in \mathbb{R}^{3 \times 3}$ where the norm of $\Delta K$ is bounded. The matrix 
\begin{equation}
K =
\begin{bmatrix}
f_x + \Delta f_x & 0 & c_x  + \Delta c_x\\
0 & f_y  + \Delta f_y &  c_y  + \Delta c_y\\
0 & 0 & 1
\end{bmatrix}
\end{equation}\label{eq:intrinsic_noise}
Note that we will denote $\mathbf{c} = [c_x, c_y]$ and $\mathbf{f} = [f_x, f_y]$ for simplicity.
Similarly, we use the extrinsics initial values $R_0$ and $\mathbf{t}_0$ and residual parameters to represent the camera rotation $R$ and translation $\mathbf{t}$. However, directly learning the rotation offset for each element of a rotation matrix would break the orthogonality of the rotation matrix. Thus, we adopt the 6-vector representation~\cite{zhou2019continuity} which uses unnormalized first two columns of a rotation matrix to represent a 3D rotation:
\begin{equation}
f\left(
\begin{bmatrix}
| & | \\
\ba_1 & \ba_2 \\
| & |
\end{bmatrix}\right) = 
\begin{bmatrix}
| & | & | \\
\bb_1 & \bb_2 & \bb_3 \\
| & | & |
\end{bmatrix},
\label{eq:6d-representation}
\end{equation}
where $\bb_1, \bb_2, \bb_3 \in \mathbb{R}^3$ are $\bb_1 = N(\ba_1)$, $\bb_2 = N(\ba_2 - (\bb_1 \cdot \ba_2) \bb_1)$, and $\bb_3 = \bb_1 \times \bb_2$, and $N(\cdot)$ denotes L2 norm. Final rotation and translation are
$$
R = f(\mathbf{a}_0 
+ \Delta \mathbf{a}), \;\; \mathbf{t} = \mathbf{t}_0 + \Delta \mathbf{t}.
$$
We use $K$ to unproject pixels to rays. The ray from the intrinsics is $\tilde{\mathbf{r}}(\mathbf{p})_d = K^{-1}\mathbf{p}$ and $\tilde{\mathbf{r}}_o = \mathbf{0}$ where $\tilde{\cdot}$ denotes a vector in the camera coordinate system. We use the extrinsics $R, \mathbf{t}$ to convert these into vectors in the world coordinate:
\begin{equation}
\mathbf{r}_d = R K^{-1}\mathbf{p}, \;\; \mathbf{r}_o = \mathbf{t}.
\label{eq:r_d}
\end{equation}
Since these ray parameters ($\mathbf{r}_d, \mathbf{r}_o$) are functions of intrinsics and extrinsics residuals ($\Delta \mathbf{f}, \Delta \mathbf{c}, \Delta \mathbf{a}, \Delta \mathbf{t}$), we can pass gradients from the rays to the residuals to optimize the parameters. Note that we do not optimize $K_0, R_0, \mathbf{t}_0$.

Cameras are made of a set of circular lenses which warp rays to the center. Thus, distortions at the edge of the lenses create circular distortion patterns. We extend our model to incorporate such radial distortions. Following the radial fisheye model in COLMAP~\cite{schoenberger2016sfm}, we adopt the fourth order radial distortion model which drops rare higher order distortions, i.e. $\mathbf{k} = (k_1 + z_{k_1}, k_2 + z_{k_2})$. 
\begin{align}
    n & = ((\mathbf{p_x} - c_x) / c_x, (\mathbf{p_y} - c_y) / c_y, 1) \\
    d & = (1 + \mathbf{k}_1n_x^2 + \mathbf{k}_2n_x^4, 1 + \mathbf{k}_1n_y^2 + \mathbf{k}_2n_y^4) \\
    \mathbf{p'} & = (\mathbf{p_x}d_x , \mathbf{p_y}d_y, 1) \\
    \mathbf{r}_d &= R K^{-1} \mathbf{p'}, \mathbf{r}_o = t
\end{align}
Similar to other camera parameters, we learn these camera parameters using photometric errors.

\subsection{Generic Non-Linear Ray Distortion}

\begin{figure}[h!]
    \includegraphics[width=1.0\columnwidth]{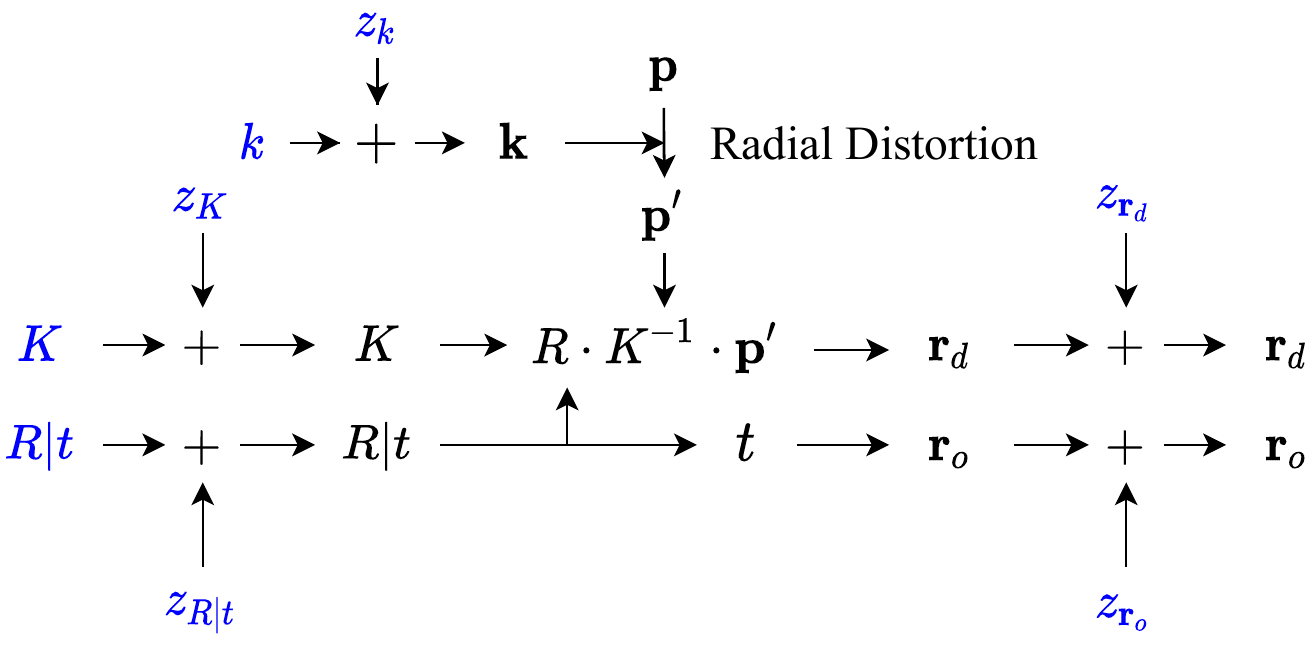}
    \vspace{0.5mm}
    \caption{Computation graph of $\mathbf{r}_o, \mathbf{r}_d$ from camera parameters and camera parameter noise. }
    \label{fig:compgraph}  
\end{figure}

We model some distortions that are easy to express mathematically. However, complex optical abberations in real lenses cannot be modeled using a parametric camera. For such noise, we use non-linear model following Grossberg~\Etal~\cite{grossberg2001general,schops2020having} to use local raxel parameters to capture generic non-linear aberration. Specifically, we use local ray parameter residuals $\mathbf{z}_d = \Delta \mathbf{r}_d(\mathbf{p}), \mathbf{z}_o = \Delta \mathbf{r}_o(\mathbf{p})$ where $\mathbf{p}$ is the image coordinate.
$$
\mathbf{r}_d' = \mathbf{r}_d + \mathbf{z}_d, \;\; \mathbf{r}_o' = \mathbf{r}_o + \mathbf{z}_o.
$$
We use bilinear interpolation to locally extract continuous ray distortion parameters
\begin{align}
    \mathbf{z}_d(\mathbf{p}) = & \sum_{x = \lfloor\mathbf{p}_x\rfloor}^{\lfloor\mathbf{p}_x\rfloor + 1} \sum_{y = \lfloor\mathbf{p}_y\rfloor}^{ \lfloor\mathbf{p}_y\rfloor + 1} (1 - |x - \mathbf{p}_x|) (1 - |y - \mathbf{p}_y|)  \nonumber\\
    & \mathbf{z}_{d}[x, y] \mathbf{z}_{d}[x, y].
\end{align}
$\mathbf{z}_{d}[x, y]$ indicates the ray direction offset at a discrete 2D coordinate $(x, y)$. We learn the parameters of $\mathbf{z}_d$ at discrete locations only. Similarly, we can define $\mathbf{z}_o(\mathbf{p})$ as bilinear interpolation of $\mathbf{z}_o[x,y]$. The final ray direction, ray offset generation can be summarized as Fig.~\ref{fig:compgraph}. 

%% file: sections/5_geometric_and_photometric.tex
\section{Geometric and Photometric Consistency}

Our camera model incorporates the generic non-linear distortions that increase the number of camera parameters drastically. In this work, we proposed using both geometric and photometric consistencies for self-calibration, which allows more accurate camera parameter calibration as these consistencies provide additional constraints. We discuss each of the constraints in this section.

\subsection{Geom. Consistency: Projected Ray Distance}
\label{sec:projected_ray_distance}
The generic camera model poses a new challenge defining a geometric loss. In most traditional work, the geometric loss is defined as an epipolar constraint that measures the distance between an epipolar line and the corresponding point, or reprojection error where a 3D point for a correspondence is defined first which is then projected to an image plane to measure the distance between the projection and the correspondence. However, these methods have few limitations when we use our generic noise model.

First, the epipolar distance assumes a perfect pinhole camera, which breaks in our setup. Second, the 3D reprojection error requires creating a 3D point cloud reconstruction using a non-differentiable process, and the camera parameters are learned indirectly from the 3D reconstruction.

In this work, rather than requiring a 3D reconstruction to compute an indirect loss like the reprojection error, we propose the projected ray distance loss that directly measures the discrepancy between rays.
Let $(\mathbf{p}_A \leftrightarrow \mathbf{p}_B)$ be a correspondence on camera 1 and 2 respectively. When all the camera parameters are calibrated, the ray $\mathbf{r}_A$ and $\mathbf{r}_B$ should intersect at the 3D point that generated point $\mathbf{p}_A$ and $\mathbf{p}_B$.

However, when there's a misalignment due to an error in camera parameters, we can measure the deviation by computing the shortest distance between corresponding rays.

\begin{figure}[ht]
    \centering
    \includegraphics[width=0.7\columnwidth]{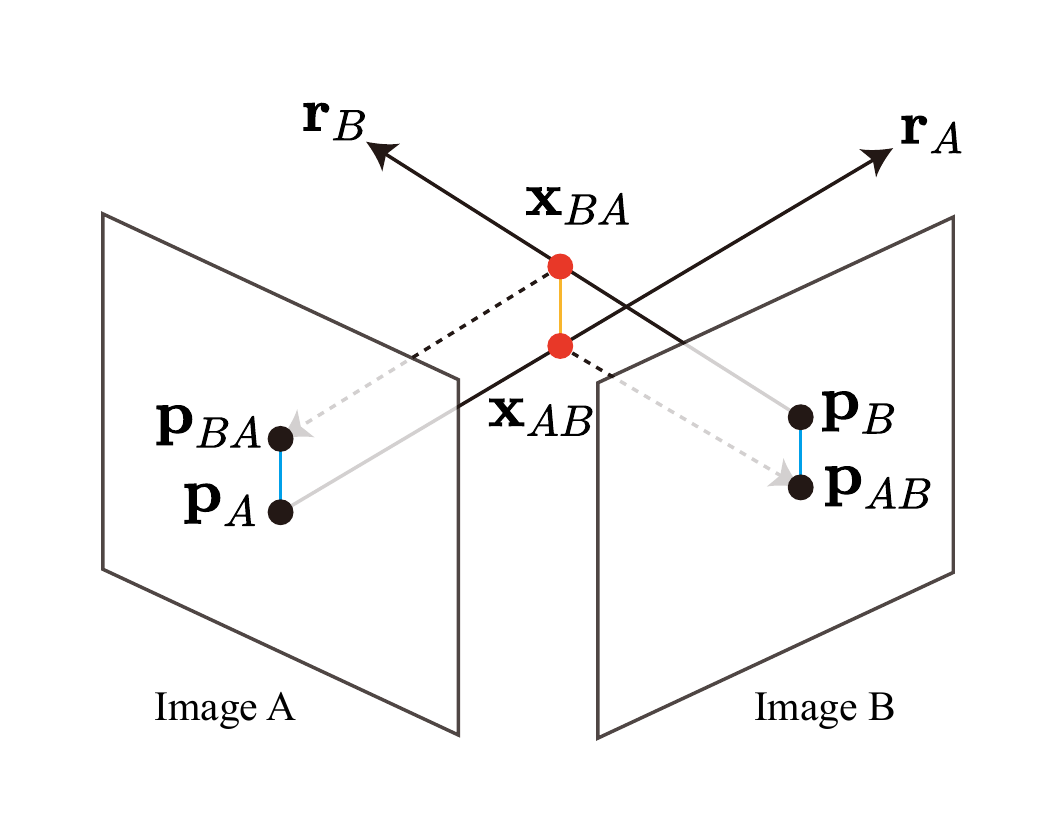}
    \vspace{0.5mm}
    \caption{An illustration of the proposed Projected Ray Distance (PRD). PRD measures the length of the projection of the shorted line segment between two rays.
    }
    \label{fig:prd}  
\end{figure}

Let a point on line A be $\mathbf{x}_{A}(t_A) = \mathbf{r}_{o, A} + t_A \mathbf{r}_{d, A}$ and a point on line B be $\mathbf{x}_{B}(t_B) = \mathbf{r}_{o, B} + t_B \mathbf{r}_{d, B}$.
A distance between the line A and a point on the line B is
\begin{equation}
d = \frac{|(\mathbf{r}_{o, B} + t_B \mathbf{r}_{d, B} - \mathbf{r}_{o,A}) \times \mathbf{r}_{A,d}|}{\mathbf{r}_{A,d} \cdot \mathbf{r}_{A,d}}
\end{equation}
If we solve for $\frac{\mathbf{d} d^2}{\mathbf{d} t_B}|_{\hat{t}_B} = 0$, we get
\begin{equation}
\hat{t}_B = \frac{(\mathbf{r}_{A,o} - \mathbf{r}_{B,o}) \times \mathbf{r}_{A,d} \cdot (\mathbf{r}_{A,d} \times \mathbf{r}_{B,d})}{(\mathbf{r}_{A,d} \times \mathbf{r}_{B,d})^2}.
\end{equation}
We substitute $\hat{t}_B$ to the line 2 and can get the $\hat{\mathbf{x}}_B = \mathbf{x}_B(\hat{t}_B)$. Similarly, we can get $\hat{\mathbf{x}}_A$. For simplicity, we will denote $\mathbf{x}_\cdot$ as $\hat{\mathbf{x}}_\cdot$ since we will focus primarily on the final solution. The distance between two points $\hat{d} = \overline{\mathbf{x}_A\mathbf{x}_B}$ is
\begin{equation}
\hat{d} = \frac{\left| (\mathbf{r}_{A,o} - \mathbf{r}_{B,o}) \cdot (\mathbf{r}_{A,d} \times \mathbf{r}_{B,d}) \right|}{|\mathbf{r}_{A,d} \times \mathbf{r}_{B,d}|}
\end{equation}
However, this distance is not normalized for correspondences. Given the same camera distortions, a correspondence for a point farther from the cameras would have a larger deviation, while a correspondence for a point closer to the cameras would have a smaller deviation. Thus, we need to normalize the scale of the distance. Thus, we project the points $\mathbf{x}_A, \mathbf{x}_B$ to image planes $I_A, I_B$ and compute distance on the image planes, rather than directly using the distance in the 3D space.
\begin{equation}
d_\pi = \frac{\|\pi_A(\mathbf{x}_B) - \mathbf{p}_A\| + \|\pi_B(\mathbf{x}_A) - \mathbf{p}_B\|}{2}
\end{equation}
where $\pi(\cdot)$ is a projection function and equalizes the contribution from each correspondence irrespective of their distance from the cameras. We visualize the projected ray distance in Fig.~\ref{fig:prd}.

This projected ray distance is a novel geometric loss different from the epipolar distance or the reprojection error. The epipolar distance is defined only for linear pinhole cameras and cannot model the non-linear camera distortions. On the other hand, the reprojection error requires extracting a 3D reconstruction in a non-differentiable preprocessing stage and optimizes the camera parameters via optimizing the 3D reconstruction. Our projected ray distance does not require the intermediate 3D reconstruction and can model the non-linear camera distortions.

\subsection{Chirality Check}

When the camera distortion is large and the baseline between cameras is small, the shortest line between rays from a correspondence might be located behind the cameras. Minimizing such invalid ray distance would result in suboptimal camera parameters. Thus, we check whether the points are behind a camera by computing the z-depth along with the camera rays. Mathematically,
\begin{equation}
R_A\mathbf{x}_B[z] > 0, \;\; R_B\mathbf{x}_A[z] > 0
\end{equation}
where $\mathbf{x}[z]$ indicates the z component of a vector.
Finally, we only average valid projected ray distances for all correspondences to compute the geometric loss.

\subsection{Photometric Consistency}

Unlike geometric consistency, photometric consistency requires reconstructing the 3D geometry because the color of a 3D point is valid only if it is visible from the current perspective. In our work, we use a neural radiance field~\cite{mildenhall2020nerf} to reconstruct the 3D occupancy and color. This implicit representation is differentiable through both position and color value and allows us to capture the visible surface through volumetric rendering. Specifically, during the rendering process, a ray is parametrized using $K_0, R_0, \mathbf{t}_0$ as well as $\Delta K, \Delta a, \Delta t$ as well as $ \mathbf{z}_o[\cdot], \mathbf{z}_d[\cdot]$ as visualized in Fig.~\ref{fig:compgraph}. We differentiate the following energy function with respect to the learnable camera parameters to optimize our self-calibration model.
\begin{equation}
    \mathcal{L} =  \sum_{\mathbf{p} \in \mathcal{I}} \abs{C(\mathbf{p}) - \hat{C}(\mathbf{r}(\mathbf{p})}_2^2
    \label{eq:nerfenergy}
\end{equation}
Here, $\mathbf{p}$ is a pixel coordinate, and $\mathcal{I}$ is a set of pixel coordinates in an image. $\hat{C}(\mathbf{r})$ is the output of the volumetric rendering using the ray $\mathbf{r}$, which corresponds to the pixel $\mathbf{p}$. $C(\mathbf{p})$ is the ground truth color. Thus, the gradient for the intrinsics is
$$
\frac{\partial L}{\partial \Delta K} = \frac{\partial L}{\partial \mathbf{r}} \left(\frac{\partial \mathbf{r}}{\partial \mathbf{r}_d} \frac{\partial \mathbf{r}_d}{\partial \Delta K}+ \frac{\partial \mathbf{r}}{\partial \mathbf{r}_o} \frac{\partial \mathbf{r}_o}{\partial \Delta K} + \frac{\partial L}{\partial{\mathbf{r}_d}}\frac{\mathbf{r}_d}{\partial \Delta k}\right).
$$
Similarly, we can define gradients for the rest of the parameters $\Delta a, \Delta t$ as well as $ \mathbf{z}_o[\cdot], \mathbf{z}_d[\cdot]$ and calibrate cameras.

%% file: sections/6_optimization.tex
\section{Optimizing Geometry and Camera}

To optimize geometry and camera parameters, we learn the neural radiance field and the camera model jointly.
However, it is impossible to learn accurate camera parameters when the geometry is unknown or too coarse for self-calibration. Thus, we sequentially learn parameters: geometry and a linear camera model first and complex camera model parameters.

\subsection{Curriculum Learning}

The camera parameters determine the positions and directions of the rays for NeRF learning, and unstable values often result in divergence or sub-optimal results. Thus, we add a subset of learning parameters to the optimization process to jointly reduce the complexity of learning cameras and geometry.
First, we learn the NeRF networks while initializing the camera focal lengths and focal centers to half the image width and height. Learning coarse geometry first is crucial since it initializes the networks to a more favorable local optimum for learning better camera parameters.
Next, we sequentially add camera parameters for the linear camera model, radial distortion, and nonlinear noise of ray direction, ray origin to the learning. We learn simpler camera models first to reduce overfitting and faster training.

\subsection{Joint Optimization}

We present the final learning algorithm in Alg.~\ref{alg:1}. The $get\_params$ function returns a set of parameters for the curriculum learning which progressively adds complexity to the camera model.
Next, we train the model with the projected ray distance by selecting a target image at random with sufficient correspondences. Heuristically, we found selecting images within maximum 30\textdegree from the source view gives an optimal result.

\begin{algorithm}
    \caption{Joint Optimization of Color Consistency Loss and Ray Distance Loss using Curriculum Learning}
    \begin{algorithmic}
    \State{Initialize NeRF parameter $\Theta$}
    \State{Initialize camera parameter $z_K$, $z_{R|t}$,  $z_{ray\_o}$, $z_{ray\_d}$, $z_{k}$}
    \State{Learnable Parameters $\mathcal{S} = \{\Theta\}$ } 
    \For {iter=1,2,...}
        \State{S' = get\_params$($iter$)$} \hfill \Comment{Curriculum learning}
        \State{$\mathbf{r}_d, \mathbf{r}_o \gets$ camera model($K, \mathbf{z}$)} \hfill Sec.~\ref{sec:camera_model}
        \State{$\mathcal{L} \gets$ volumetric rendering($\mathbf{r}_d, \mathbf{r}_o, \Theta$)} \hfill Eq.~\ref{eq:nerf_volumetric_rendering}
        \If {iter \% $n$ == 0 and iter $>= n_{prd}$}
            \State{$I' \gets $ random($R_I, t_I, \mathcal{I}$)}
            \State{$\mathcal{C} \gets$ Correspondence($I$, $I'$)}
            \State{$\mathcal{L}_{prd}\gets$ Projected Ray Distance($\mathcal{C}$}) \hfill Sec.~\ref{sec:projected_ray_distance}
            \State{$\mathcal{L} \gets \mathcal{L} + \lambda\mathcal{L}_{prd}$}
        \EndIf
        \For{$s \in \mathcal{S'}$}
            \State{$\mathbf{s} \gets \mathbf{s} + \nabla_{\mathbf{s}} \mathcal{L}$}
        \EndFor
    \EndFor
    \end{algorithmic}
    \label{alg:1}
\end{algorithm}

%% file: sections/7_experiment.tex
\section{Experiment}

\subsection{Dataset}

We use three datasets to analyze different aspects of our model. 
Two outdoor scenes, Mildenhall~\etal~\cite{mildenhall2019local} and Zhang~\etal~\cite{knapitsch2017tanks}, are captured with a pinhole camera lens. LLFF~\cite{mildenhall2019local} and Tanks and Temples dataset~\cite{knapitsch2017tanks} are composed of 8 and 4 scenes, respectively, where their camera parameters are estimated using COLMAP~\cite{schoenberger2016sfm}. 

Since these datasets are captured using professional cameras with small lens distortions, we collected a few scenes using a fish-eye camera to examine the end-to-end learning capacity of our model. We acquire the camera information with COLMAP. 

\subsection{Self-Calibration}
\begin{figure*}
    \centering
    \includegraphics[width=1.00\textwidth]{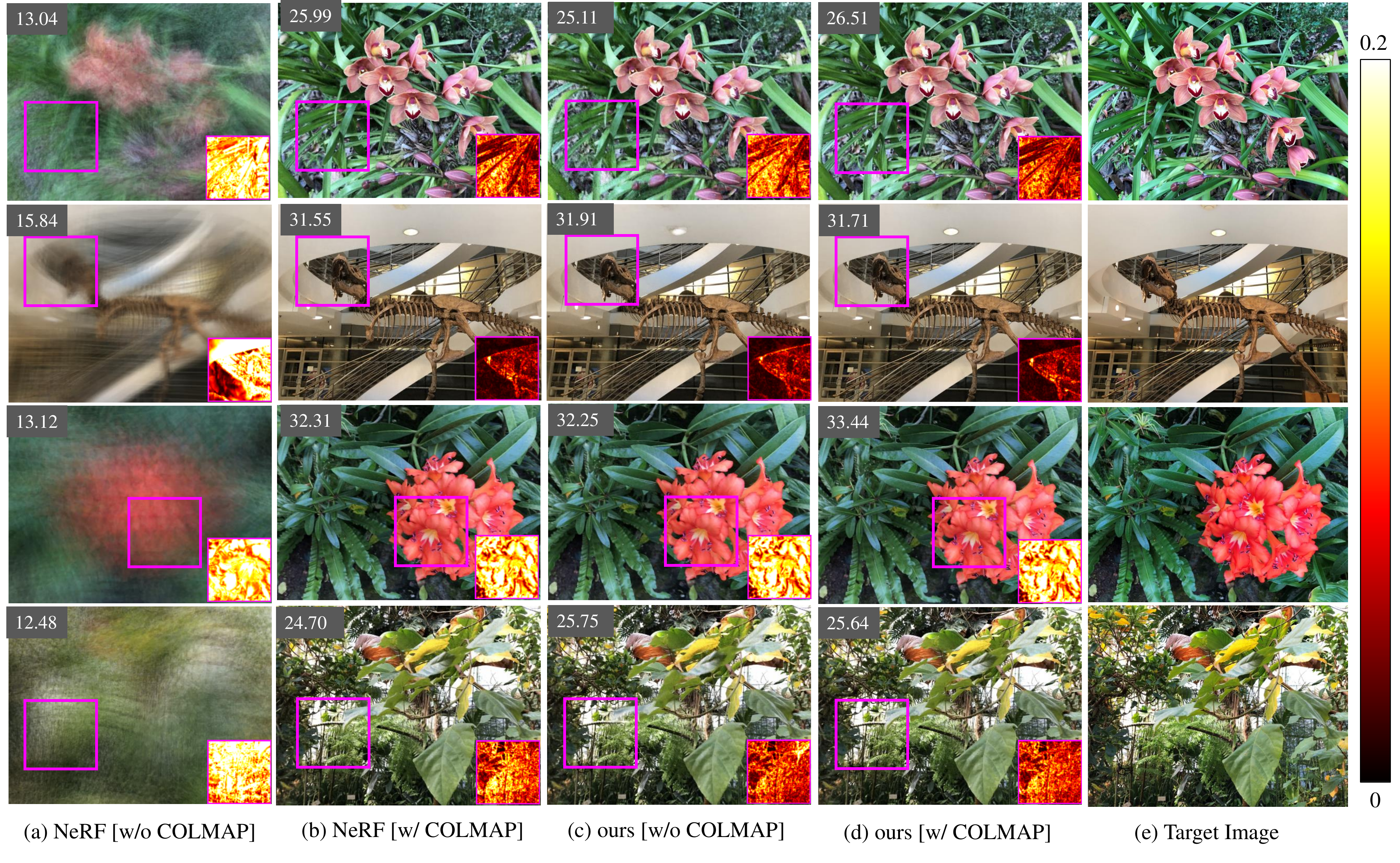}
    \caption{Comparison of NeRF~\cite{mildenhall2020nerf} and our approach using LLFF~\cite{mildenhall2019local} dataset. The first two columns of images show results of NeRF without or with intrinsic and extrinsic camera parameters. We use COLMAP~\cite{schonberger2016structure} for the camera initialization. The third and fourth columns show rendered images using our approach with the same configuration. Our self-calibration approach shows consistent results regardless of using the camera prior. For each subfigure, PSNR is shown on the upper left. 
    \label{fig:self_calibration}
    }
\end{figure*}

We train our model from scratch to demonstrate that our model can self-calibrate the camera information. We initialize all the rotation matrices, the translation vectors, and focal lengths to an identity matrix, zero vector, and height and width of the captured images. Table~\ref{tab:nerf_wo_colmap} reports the qualities of the rendered images in the training set. Although our model does not adopt calibrated camera information, our model shows a reliable rendering performance. Moreover, for some scenes, our model outperforms NeRF, trained with COLMAP~\cite{schoenberger2016sfm} camera information. We have visualized the rendered images in Figure~\ref{fig:self_calibration}.

\begin{table}[ht]
\caption{
Comparison of NeRF and our model when no calibrated camera information is given. 
"nan" denotes the case when no inlier matches are acquired due to the wrong camera information.
\label{tab:nerf_wo_colmap}
}
\vspace{3mm}
\centering
\resizebox{.46\textwidth}{!}{

\begin{tabular}{c|c|c}
Scene  & Model & PSNR($\uparrow$) / SSIM($\uparrow$) / LPIPS($\downarrow$) / PRD($\downarrow$)               \\ 
\hline
\multirow{2}{*}{Flower}  & NeRF & 13.8 / 0.302 / 0.716 / nan  \\
                          & ours  &\textbf{33.2 / 0.945 / 0.060 / 0.911} \\ 
\hline
\multirow{2}{*}{Fortress}     & NeRF  &  16.3 / 0.524 / 0.445 / nan \\
                        & ours  &  \textbf{35.7 / 0.945 / 0.069 / 0.833} \\ 
\hline
\multirow{2}{*}{Leaves}   & NeRF  & 13.01 / 0.180 / 0.687 / nan  \\
                          & ours  & \textbf{ 25.75 / 0.878 / 0.146 / 0.885} \\ 
\hline
\multirow{2}{*}{Trex}      & NeRF  & 15.70 / 0.409 / 0.575 / nan \\
                          & ours  & \textbf{31.75 / 0.954 / 0.104 / 1.002}
\end{tabular}
}

\end{table}

\subsection{Improvement over NeRF}
\begin{figure}[ht]
\centering
    \includegraphics[width=0.97\columnwidth]{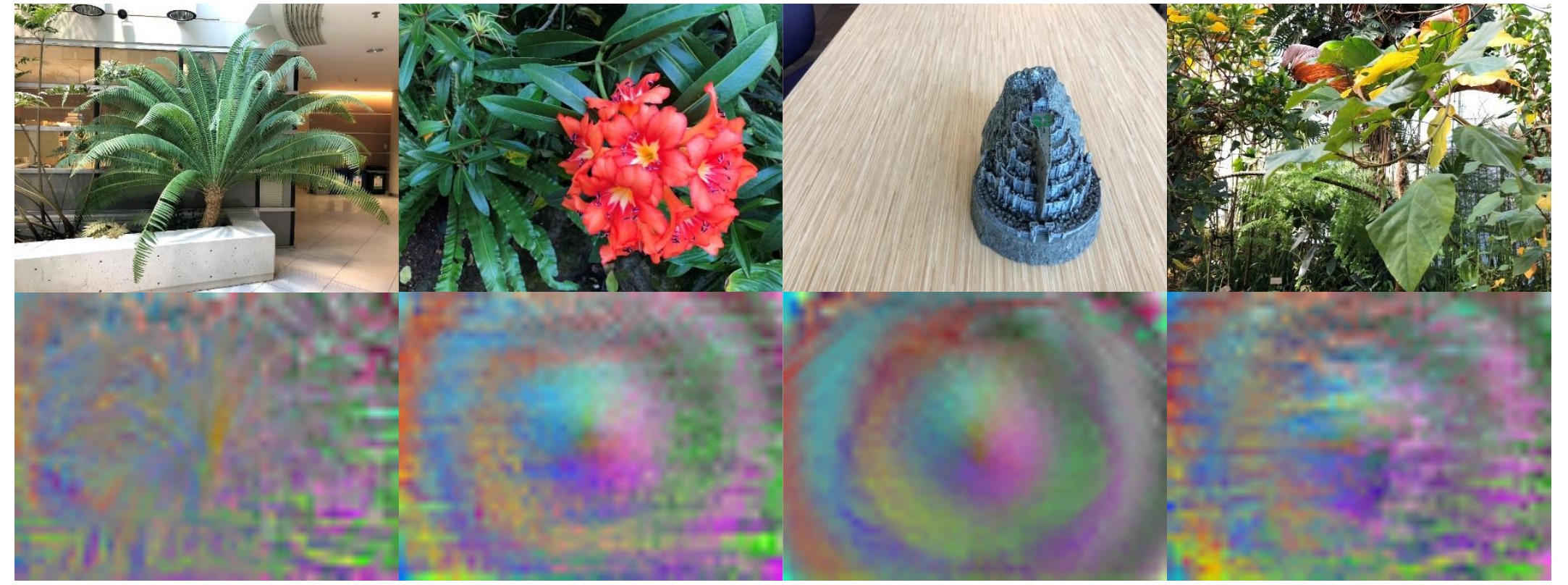}
    \vspace{1.5mm}
    \caption{Visualization of the captured non-linear distortions. The second row shows learned ray offsets.
    }
    \label{fig:qual_distortion}
\end{figure}

We have observed that our model shows better rendering qualities than NeRF when COLMAP initializes the camera information. We compare the rendering qualities of NeRF and our model in Table~\ref{tab:nerf_colmap}. Our model consistently shows better rendering qualities than the original NeRF. In addition, our model indicates much less projected ray distance, indicating that our model improves the camera information. We visualize the non-linear distortion that our camera model learned in Fig.~\ref{fig:qual_distortion}.

\begin{table}[ht]
\caption{
Comparison of NeRF and our model when the camera parameters are initialized with COLMAP~\cite{schoenberger2016sfm} in LLFF~\cite{mildenhall2019local} dataset.
\label{tab:nerf_colmap}
}
\centering
\resizebox{.46\textwidth}{!}{

\begin{tabular}{c|c|c}
Scene       & Model & PSNR($\uparrow$) / SSIM($\uparrow$) / LPIPS($\downarrow$) / PRD($\downarrow$)               \\ 
\hline
\multirow{2}{*}{Flower}  & NeRF  &  32.2 / 0.937 / 0.067 / 2.440  \\
                          & ours  & \textbf{33.3 / 0.946 / 0.058 / 0.895} \\ 
\hline
\multirow{2}{*}{Fortress}     & NeRF  & 35.3 / 0.947 / 0.056 / 2.475 \\
                          & ours  & \textbf{36.6 / 0.960 / 0.049 / 0.724}  \\ 
\hline
\multirow{2}{*}{Leaves}   & NeRF  & 25.3 / 0.874 / 0.149 / 2.709 \\
                          & ours  & \textbf{25.9 / 0.886 / 0.136 / 0.85}4 \\ 
\hline
\multirow{2}{*}{Trex} & NeRF  & 31.4 / 0.955 / 0.099 / 2.368 \\
                          & ours  &  \textbf{32.0 / 0.959 / 0.095 / 0.953}
\end{tabular}

}
\end{table}

\subsection{Improvement over NeRF++}
\begin{figure*}
    \centering
    \includegraphics[width=1.00\textwidth]{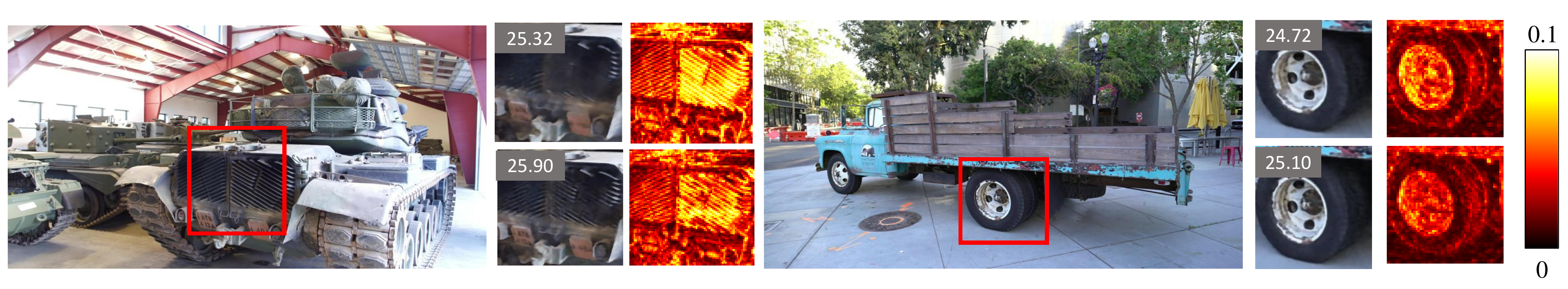}
    \caption{Experiments using tanks and temples~\cite{knapitsch2017tanks} dataset. For each scene, the zoom-in of rendered images and error maps (0 to 0.1 pixel intensity range) are presented, and they are obtained from NeRF++~\cite{zhang2020nerf++} (first row) and our model (second row). For each subfigure, PSNR is shown on the upper left. \label{fig:qual_nerfpp}
    \vspace{2mm}
    }
\end{figure*}

Since our model is designed to work on variants of NeRF, we have substituted the NeRF architecture to the NeRF++~\cite{zhang2020nerf++} architecture. We then compare NeRF++ and our model in tanks and temples~\cite{knapitsch2017tanks} dataset. Table~\ref{tab:npp_real} reports rendering qualities and projected ray distance loss in the training set. Our model results in better rendering qualities and much less train projected ray distance. The qualitative results are visualized in Figure~\ref{fig:qual_nerfpp}. 

\begin{table}[ht]
\caption{Rendering qualities of NeRF++ and our model in tanks and temples~\cite{knapitsch2017tanks} dataset. 
\label{tab:npp_real}}

\vspace{2mm}
\centering
\resizebox{0.47\textwidth}{!}{
\begin{tabular}{c|c|c}
Scene                       & Model  & PSNR($\uparrow$) / SSIM($\uparrow$) / LPIPS($\downarrow$) / PRD($\downarrow$) \\ 
\hline
\multirow{2}{*}{M60}       & NeRF++ & 25.62 / 0.772 / 0.395 / 1.335 \\
                            & ours   & \textbf{26.99 / 0.805 / 0.359 / 1.326}  \\ 
\hline
\multirow{2}{*}{Playground}      & NeRF++ &  25.14 / 0.681 / 0.434 / 1.302 \\
                            & ours   & \textbf{26.17 / 0.715 / 0.396 / 1.299}   \\ 
\hline
\multirow{2}{*}{Train} & NeRF++ & 21.80 / 0.619 / 0.479 / 1.261 \\
                            & ours   & \textbf{22.71 / 0.651 / 0.450 / 1.255} \\ 
\hline
\multirow{2}{*}{Truck}        & NeRF++ & 24.13 / 0.730 / 0.392 / 1.248 \\
                            & ours   & \textbf{25.22 / 0.763 / 0.352 / 1.240} 
\end{tabular}
}

\end{table}

\subsection{Fish-eye Lens Reconstruction}

We test our model on images with high distortion to contrast the importance of end-to-end learning of camera parameters.
Conventional feature matching algorithms fail to acquire reliable correspondences for these scenes, so we skip the projected ray distance loss from our curriculum training. Table~\ref{tab:fisheye} reports the rendering qualities of our learned model and the baseline NeRF++. We trained the baseline and our model from the COLMAP initialization with a radial distortion model that provides fish-eye camera parameters. Since the NeRF++ camera model does not incorporate the radial distortion, we modify the implementation to incorporate the fish-eye distortion in ray computation.

\begin{figure}[ht]
    \centering
    \includegraphics[width=0.47\textwidth]{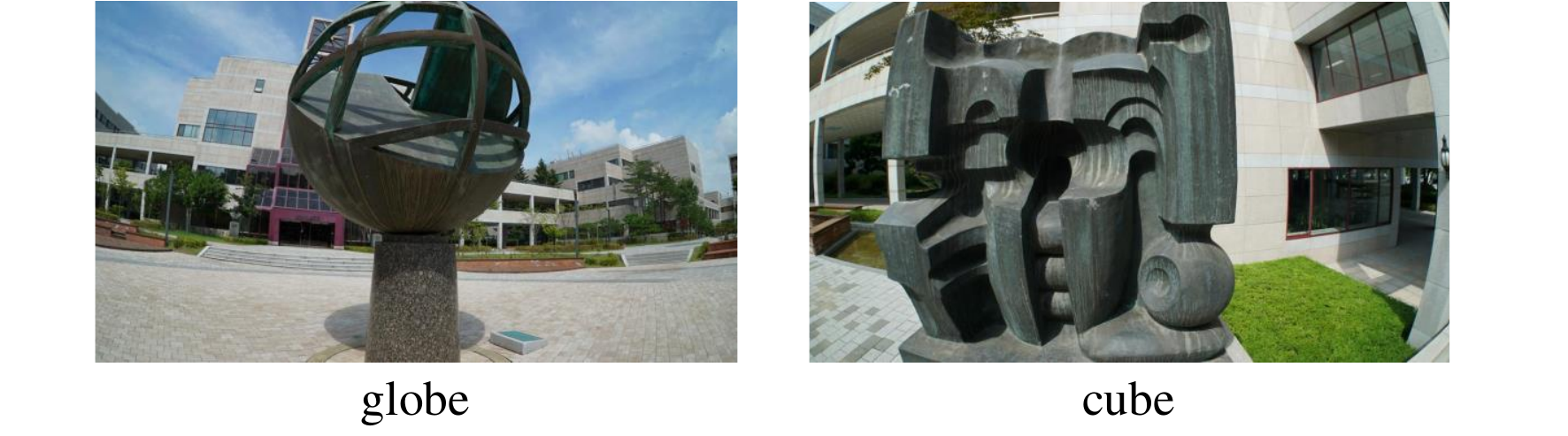}
    \vspace{1.5mm}
    \caption{Images captured using a fish-eye camera. 
    \label{fig:target_fisheye}
    }
\end{figure}

\begin{table}
\caption{Rendering qualities of scenes captured on fish-eye cameras. "RD" denotes the modified implementation to reflect radial distortions. \label{tab:fisheye}}
\vspace{3mm}
\centering
\resizebox{0.9\columnwidth}{!}{
\begin{tabular}{c|c|c}
Scene                   & Model & PSNR($\uparrow$) / SSIM($\uparrow$) / LPIPS($\downarrow$) \\ 
\hline
\multirow{2}{*}{Globe} & NeRF++[RD]  &  21.97 / 0.572 / 0.659  \\
                        & ours  & \textbf{23.76 / 0.598 / 0.633}  \\ 
\hline
\multirow{2}{*}{Cube} & NeRF++[RD] & 21.30 / 0.574 / 0.643 \\
                        & ours  &  \textbf{23.17 / 0.605 / 0.616} \\
\end{tabular}
}

\end{table}

\subsection{Ablation Study}
\begin{figure}
    \centering
    \vspace{-6mm}
    \includegraphics[width=0.75\columnwidth]{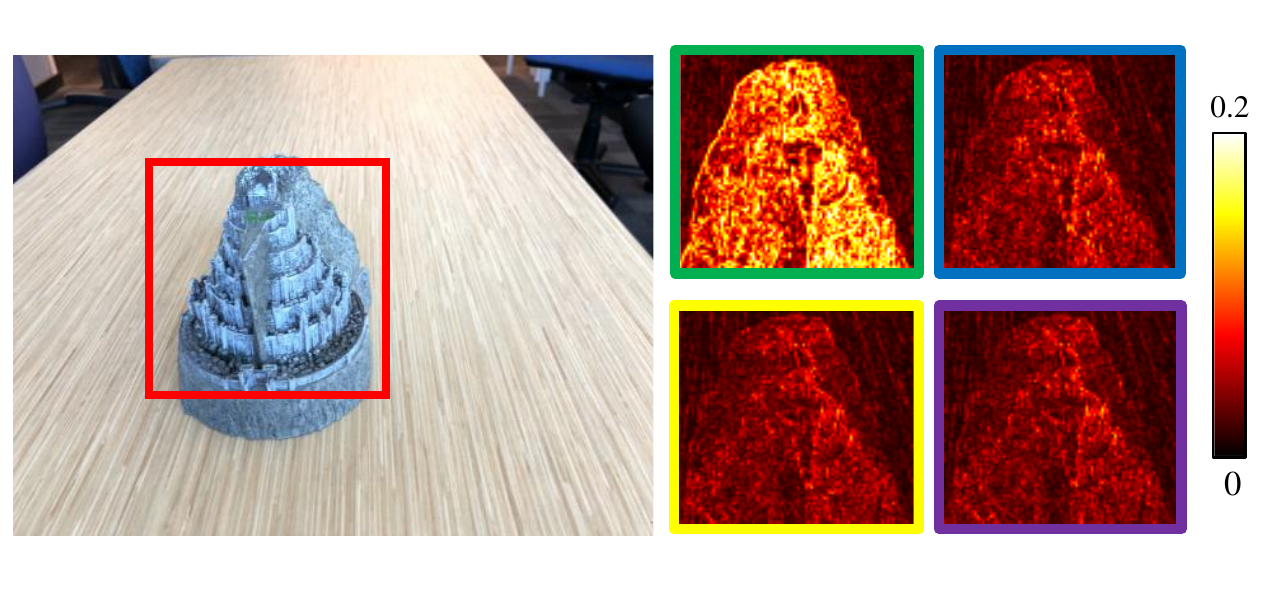}
    \vspace{-2mm}
    \caption{Visualization of rendered images for each configurations shown in Table~\ref{tab:nerf_ablation}. The green, blue, yellow, and purple box are the error map of NeRF, NeRF + IE, NeRF + IE + OD, and NeRF + IE + OD + PRD, respectively. }
    \label{fig:main_ablation}
\end{figure}

To check the effects of the proposed models, we conduct an ablation study. We check the performance for each phase in curriculum learning. We train 200K iterations for each phase. From this experiment, we have observed that extending our model is more potential in rendering clearer images. However, for some scenes, adopting projected ray distance increases the overall projected ray distance. Table ~\ref{tab:nerf_ablation} reports the results of the ablation study and Figure~\ref{fig:main_ablation} visualizes the errors.  

\begin{table}
\centering
\caption{Ablation studies about components of our model. "IE", "OD", and "PRD" denote learnable intrinsic and extrinsic parameters, learnable non-linear distortion, and projected ray distance loss, respectively. 
\label{tab:nerf_ablation}}
\vspace{3mm}
\resizebox{0.47\textwidth}{!}{
\begin{tabular}{c|c|c}
Scene                     &             & PSNR($\uparrow$) / SSIM($\uparrow$) / LPIPS($\downarrow$) / PRD($\downarrow$)               \\
\hline
\multirow{2}{*}{Fortress}     & NeRF   & 30.5 / 0.866 / 0.096 / 0.856           \\
                          & + IE & 35.3 / 0.948 / 0.058 / 0.729   \\ 
                          & + IE + OD & 36.4 / 0.957 / 0.051 / \textbf{0.724} \\
                          & + IE + OD + PRD & \textbf{36.6 / 0.96 / 0.049 / 0.724} \\
\hline
\multirow{2}{*}{Room}     & NeRF   & 31.5 / 0.950 / 0.096 / 0.883           \\
                          & + IE & 38.3 / 0.978 / 0.070 / 0.806  \\ 
                          & + IE + OD & 39.4 / 0.980 / 0.065 / 0.805 \\
                          & + IE + OD + PRD & \textbf{39.7 / 0.981 / 0.063 / 0.805} \\
\end{tabular}
}

\end{table}

%% file: sections/conclusion.tex
\vspace{-1mm}
\section{Conclusion}
\vspace{-1mm}
We propose a self-calibration algorithm that learns geometry and camera parameters jointly end-to-end. The camera model consists of a pinhole model, radial distortion, and non-linear distortion, which capture real noises in lenses.
We also propose projected ray distance to improve accuracy, which allows our model to learn fine-grained correspondences. We show that our model learns geometry and camera parameters from scratch when the poses are not given, and our model improves both NeRF and NeRF++ to be more robust when camera poses are given.

\vspace{2mm}
\noindent\textbf{Acknowledgements.}
This work was supported by the IITP grants (2019-0-01906: AI Grad. School Prog. - POSTECH and 2021-0-00537: visual common sense through self-supervised learning for restoration of invisible parts in images) funded by Ministry of Science and ICT, Korea.

%% file: supplementary/implementation_details.tex
\section{Implementation Details}
\subsection{Training NeRF Networks}
We use batch size of 1024 rays for NeRF~\cite{mildenhall2020nerf}, 512 rays for NeRF++~\cite{zhang2020nerf++}. We initially set the learning rate of NeRF to 0.0005. The learning rate decays exponentially to one-tenth for every 400000 steps. For NeRF++, we initially set the learning rate to 0.0005. The learning rate decays exponentially to one-tenth for every 7500000 steps. As explained in the main paper, we adopt curriculum learning for better stability of training. We extend our learnable camera parameters for every 200K iterations for NeRF experiments. For NeRF++, we have extended our learnable parameters in 500K iterations, 800K iterations, and 1.1M iterations. For NeRF, we use 64 samples for the coarse network and 128 samples for the fine network. In NeRF++ experiments, we have scaled extrinsic noises to 0.01. Especially for tanks and temples~\cite{knapitsch2017tanks} dataset, 64 points along a ray are sampled and fed to a coarse network. 128 points along the same ray are sampled and fed to a fine network. For FishEyeNeRF experiments, we have sampled 128 points for the coarse network and 256 for the fine network along the ray. Besides, 1024 rays are used for the FishEyeNeRF experiments for each iteration.

\subsection{Projected Ray Distance Evaluation}
The projected ray distance measures the deviation of a correspondence pair. However, as it only finds the shortest distance between rays in 3D space, a small change in the direction sometimes leads to a large change in the ray distance. Thus, we threshold ray distance with $\eta$ and remove pairs above this threshold. We set the $\eta$ to 5.0 for all the experiments. 

%% file: supplementary/llff_with_colmap.tex
\section{Calibration with COLMAP initialization}

We extend Table 2 in the main paper by conducting experiments in other scenes of the LLFF dataset~\cite{mildenhall2019local}. Table~\ref{tab:nerf_with_colmap} reports the rendering qualities and projected ray distance of NeRF and our model. Our model shows a consistent improvement from NeRF when learnable camera parameters are initialized by COLMAP camera information. 

\begin{table}[h]
\centering
\caption{Comparison between NeRF and NeRF + ours when the camera information is initialized with COLMAP. We report PSNR, SSIM, LPIPS, and PRD for training dataset.}
\label{tab:nerf_with_colmap}
\resizebox{0.46\textwidth}{!}{
\begin{tabular}{c|c|c}
scene                     &             & PSNR($\uparrow$) / SSIM($\uparrow$) / LPIPS($\downarrow$) / PRD($\downarrow$)               \\ 
\hline
\multirow{2}{*}{Fern}     & NeRF        & 30.7 / 0.912 / 0.127 / 2.369         \\
                          & ours & \textbf{31.1 / 0.917 / 0.117 / 0.993}  \\ 
\hline
\multirow{2}{*}{Flower}   & NeRF        & 32.2 / 0.937 / 0.067 / 2.440           \\
                          & ours & \textbf{33.3 / 0.946 / 0.058 / 0.895}  \\ 
\hline
\multirow{2}{*}{Fortress} & NeRF        & 35.3 / 0.947 / 0.056 / 2.475           \\
                          & ours & \textbf{36.6 / 0.96 / 0.049 / 0.724}  \\ 
\hline
\multirow{2}{*}{Horns}   & NeRF        & 31.6 / 0.931 / 0.116 / 2.499         \\
                          & ours & \textbf{32.2 / 0.932 / 0.114 / 0.907}  \\ 
\hline
\multirow{2}{*}{Leaves}     & NeRF        & 25.3 / 0.874 / 0.149 / 2.709 \\
                          & ours &   \textbf{25.9 / 0.886 / 0.136 / 0.854}         \\ 
\hline
\multirow{2}{*}{Orchids}  & NeRF        &  25.6 / 0.864 / 0.151 / 2.417          \\
                          & ours & \textbf{ 26.4 / 0.881 / 0.134 / 1.173} \\ 
\hline
\multirow{2}{*}{Room}    & NeRF        &\textbf{ 39.7 / 0.981 / 0.063 }/ 2.531 \\
                          & ours &   \textbf{39.7 / 0.981 / 0.063 / 0.805} \\ 
\hline
\multirow{2}{*}{Trex}    & NeRF        & 31.4 / 0.955 / 0.099 / 2.368 \\
                          & ours & \textbf{32.0 / 0.959 / 0.095 / 0.953} \\  
\end{tabular}
}
\end{table}

%% file: supplementary/llff_wo_colmap.tex
\section{Calibration without COLMAP initialization}
We also extend Table 1 in the main paper by conducting the experiments in other scenes of the LLFF dataset~\cite{mildenhall2019local}. Table~\ref{tab:nerf_wo_colmap_supp} reports the rendering qualities and projected ray distance metric. NeRF fails to render the scenes reliably; however,  our model does. Qualitative results are shown in Figure~\ref{fig:wo_colmap}.

\begin{table*}[h]
\centering
\caption{Comparison between NeRF and NeRF + ours when the camera information is initialized with COLMAP. We report PSNR, SSIM, LPIPS, and PRD for training dataset.}
\label{tab:nerf_wo_colmap_supp}
\resizebox{0.5\textwidth}{!}{
\begin{tabular}{c|c|c}
scene                     &             & PSNR($\uparrow$) / SSIM($\uparrow$) / LPIPS($\downarrow$) / PRD($\downarrow$)               \\ 
\hline
\multirow{2}{*}{fern}     & NeRF        & 16.9 / 0.435 / 0.544 / nan           \\
                          & ours &  \textbf{31.2 / 0.918 / 0.117 / 1.020} \\ 
\hline
\multirow{2}{*}{flower}   & NeRF        & 13.8 / 0.302 / 0.716 / nan           \\
                          & ours &  \textbf{33.2 / 0.945 / 0.060 / 0.911} \\ 
\hline
\multirow{2}{*}{fortress} & NeRF        & 16.3 / 0.524 / 0.445 / nan           \\
                          & ours & \textbf{ 35.7 / 0.945 / 0.069 / 0.833} \\ 
\hline
\multirow{2}{*}{horns}    & NeRF        & 14.8 / 0.390 / 0.634 / nan          \\
                          & ours &  \textbf{22.6 / 0.0613 / 0.494 / 1.578} \\
\hline
\multirow{2}{*}{leaves}   & NeRF        & 13.0 / 0.170 / 0.687 / nan          \\
                          & ours &  \textbf{25.8 / 0.878 / 0.146 / 0.885} \\ 
\hline
\multirow{2}{*}{orchids}  & NeRF        & 13.1 / 0.170 / 0.674 / nan           \\
                          & ours & \textbf{ 24.8 / 0.830 / 0.204 / 1.269} \\ 
\hline
\multirow{2}{*}{room}     & NeRF        & 18.1 / 0.660 / 0.486 / nan  \\
                          & ours &  \textbf{37.5 / 0.967 / 0.103 / 0.852} \\ 
\hline
\multirow{2}{*}{trex}    & NeRF        & 15.7 / 0.409 / 0.575 / nan           \\
                          & ours & \textbf{31.8 / 0.954 / 0.104 / 1.002} \\ 

\end{tabular}
}
\end{table*}

%% file: supplementary/npp_qual.tex
\begin{figure*}
    \centering
    \includegraphics[width=2.0\columnwidth]{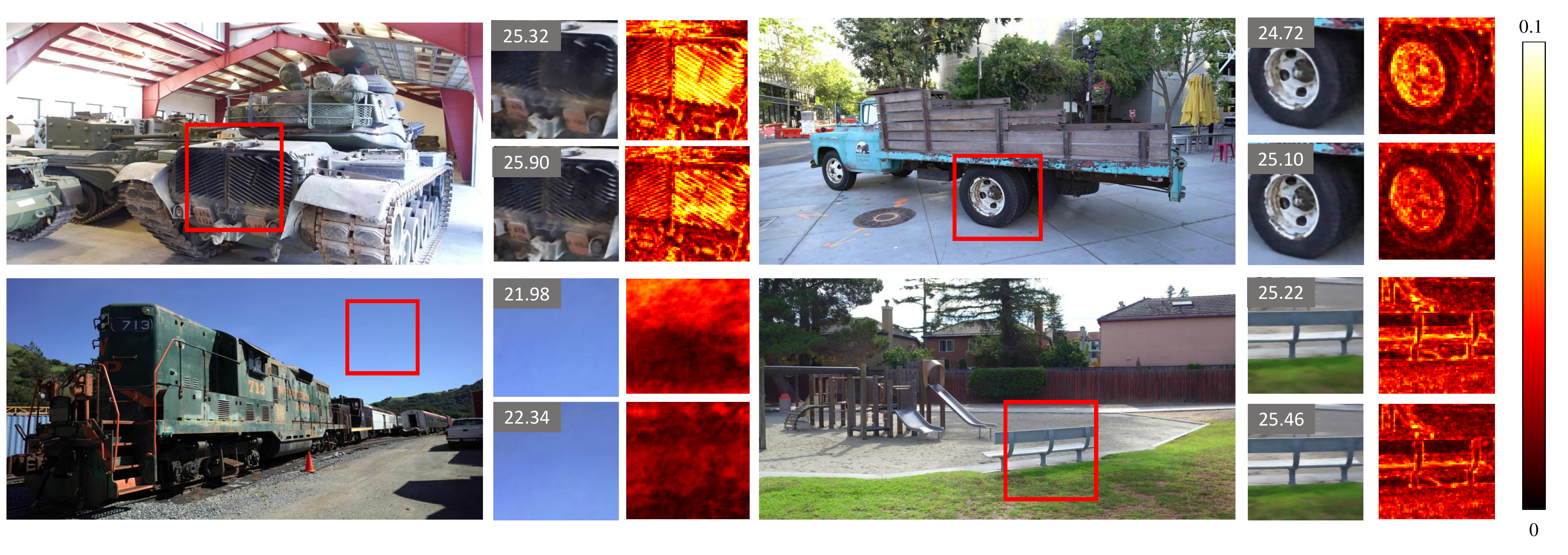}
    \vspace{-3mm}
    \caption{Error map of rendered images by NeRF++~\cite{zhang2020nerf++} and our model in tanks and temples~\cite{knapitsch2017tanks} dataset. The above and the below maps are generated error maps by NeRF++ and our model, respectively. For each subfigure, PSNR is shown on the upper left. }
    \label{fig:npp_tanks}
\end{figure*}

%% file: supplementary/npp_fisheye.tex
\begin{figure*}
    \centering
    \includegraphics[width=1.5\columnwidth]{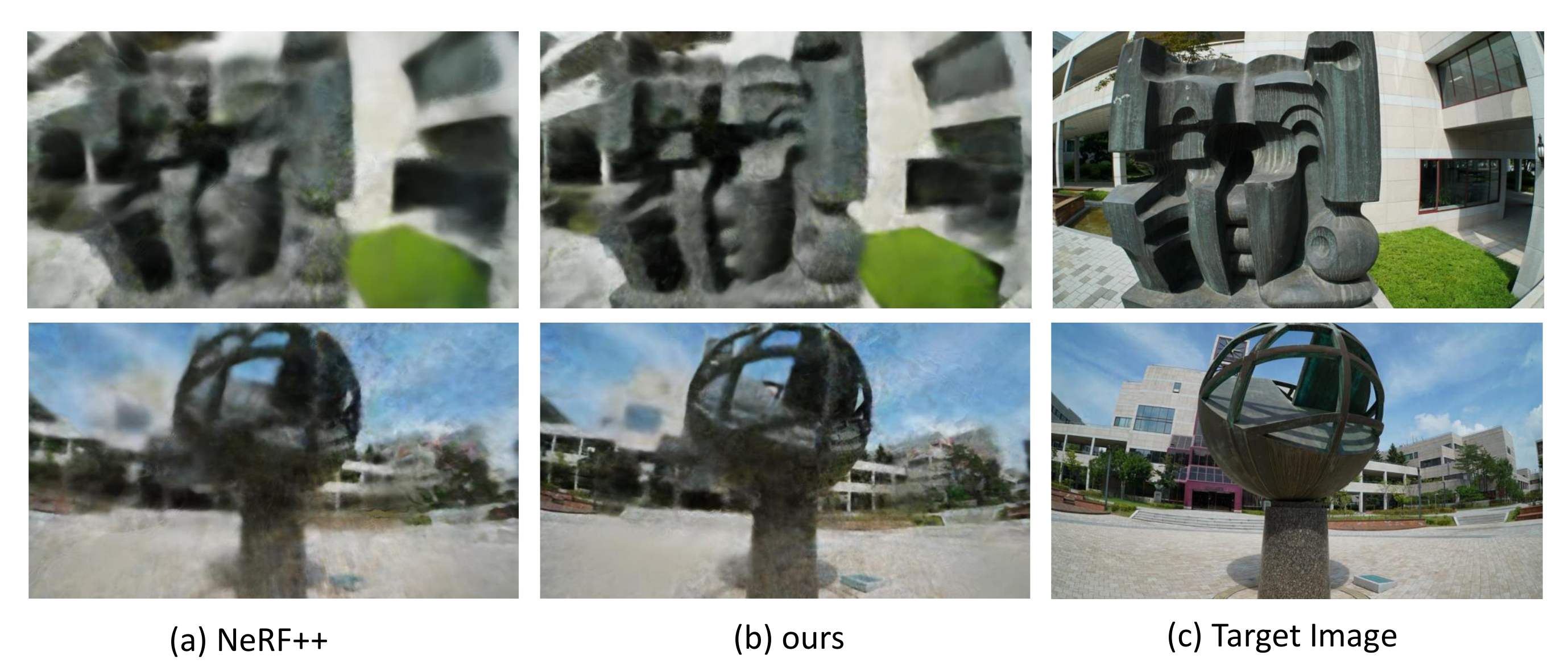}
    \caption{Comparison between NeRF++~\cite{zhang2020nerf++} and our model in FishEyeNeRF dataset. Our model shows much clearer rendering compared to NeRF++.  }
    \label{fig:fisheye_render}
\end{figure*}

%% file: supplementary/llff_ablation.tex
\section{Ablation Studies}

We extend ablation study in our main paper by conducting experiments in the other scenes of LLFF~\cite{mildenhall2019local} dataset. Table~\ref{tab:ablation} reports the quantitative results of the ablation study. 

\begin{table*}[h]
\centering
\caption{Ablation studies about components of our model. "IE", "OD", and "PRD" denote learnable intrinsic and extrinsic parameters, learnable non-linear distortion, and projected ray distance loss, respectively. }
\label{tab:ablation}
\resizebox{0.6\textwidth}{!}{
\begin{tabular}{c|c|c}
scene                     &             & PSNR($\uparrow$) / SSIM($\uparrow$) / LPIPS($\downarrow$) / PRD($\downarrow$)               \\ 
\hline
\multirow{2}{*}{Fern}     & NeRF  & 25.3 / 0.809 / 0.178 / 1.069 \\
                          & + IE & 30.2 / 0.907 / 0.127 / \textbf{0.988} \\ 
                          & + IE + OD & 30.9 / 0.915 / 0.118 / 0.991 \\
                          & + IE + OD + PRD & \textbf{31.1 / 0.917 / 0.117} / 0.993 \\
\hline
\multirow{2}{*}{Flower}     & NeRF & 28.1 / 0.879 / 0.104 / 0.989 \\
                          & + IE &  32.2 / 0.937 / 0.068 / \textbf{0.893} \\ 
                          & + IE + OD & 33.1 / 0.944 / 0.060 /\textbf{ 0.893} \\
                          & + IE + OD + PRD &\textbf{ 33.3 / 0.946 / 0.058} / 0.895 \\
\hline
\multirow{2}{*}{Fortress}     & NeRF & 30.5 / 0.866 / 0.096 / 0.856 \\
                          & + IE & 35.3 / 0.948 / 0.058 / 0.729  \\ 
                          & + IE + OD & 36.4 / 0.957 / 0.051 / \textbf{0.724}  \\
                          & + IE + OD + PRD & \textbf{36.6 / 0.960 / 0.049 / 0.724} \\
\hline
\multirow{2}{*}{Horns}     & NeRF & 27.0 / 0.857 / 0.171 / 0.987 \\
                          & + IE & 31.2 / 0.921 / 0.128 / \textbf{0.907} \\ 
                          & + IE + OD & 32.0 / 0.930 / 0.117 /\textbf{ 0.907} \\
                          & + IE + OD + PRD & \textbf{32.2 / 0.932 / 0.114 / 0.907} \\
\hline
\multirow{2}{*}{Leaves}     & NeRF & 22.0 / 0.787 / 0.193 / 0.951 \\
                          & + IE & 25.2 / 0.872 / 0.147 / 0.853 \\ 
                          & + IE + OD & 25.8 / 0.883 / 0.138 / \textbf{0.852} \\
                          & + IE + OD + PRD & \textbf{25.9 / 0.886 / 0.136 } / 0.854 \\
\hline
\multirow{2}{*}{Orchids}     & NeRF & 22.8 / 0.783 / 0.199 / 1.240 \\
                          & + IE & 25.7 / 0.866 / 0.147 / \textbf{1.170} \\ 
                          & + IE + OD & 26.3 / 0.878 / 0.137 / 1.172 \\
                          & + IE + OD + PRD & \textbf{26.4 / 0.881 / 0.134} / 1.173 \\
\hline
\multirow{2}{*}{Room}     & NeRF & 31.5 / 0.950 / 0.096 / 0.883 \\
                          & + IE & 38.3 / 0.978 / 0.070 / 0.806 \\ 
                          & + IE + OD & 39.4 / 0.980 / 0.065 / \textbf{0.805} \\
                          & + IE + OD + PRD & \textbf{39.7 / 0.981 / 0.063 / 0.805} \\
\hline
\multirow{2}{*}{Trex}     & NeRF & 26.5 / 0.893 / 0.138 / 1.016 \\
                          & + IE & 31.0 / 0.952 / 0.104 / \textbf{0.951}  \\ 
                          & + IE + OD & 31.8 / 0.958 / 0.097 / 0.952 \\
                          & + IE + OD + PRD & \textbf{32.0 / 0.959 / 0.095} / 0.953 \\

\end{tabular}
}
\end{table*}

\begin{figure*}
    \centering
    \includegraphics[width=2.0\columnwidth]{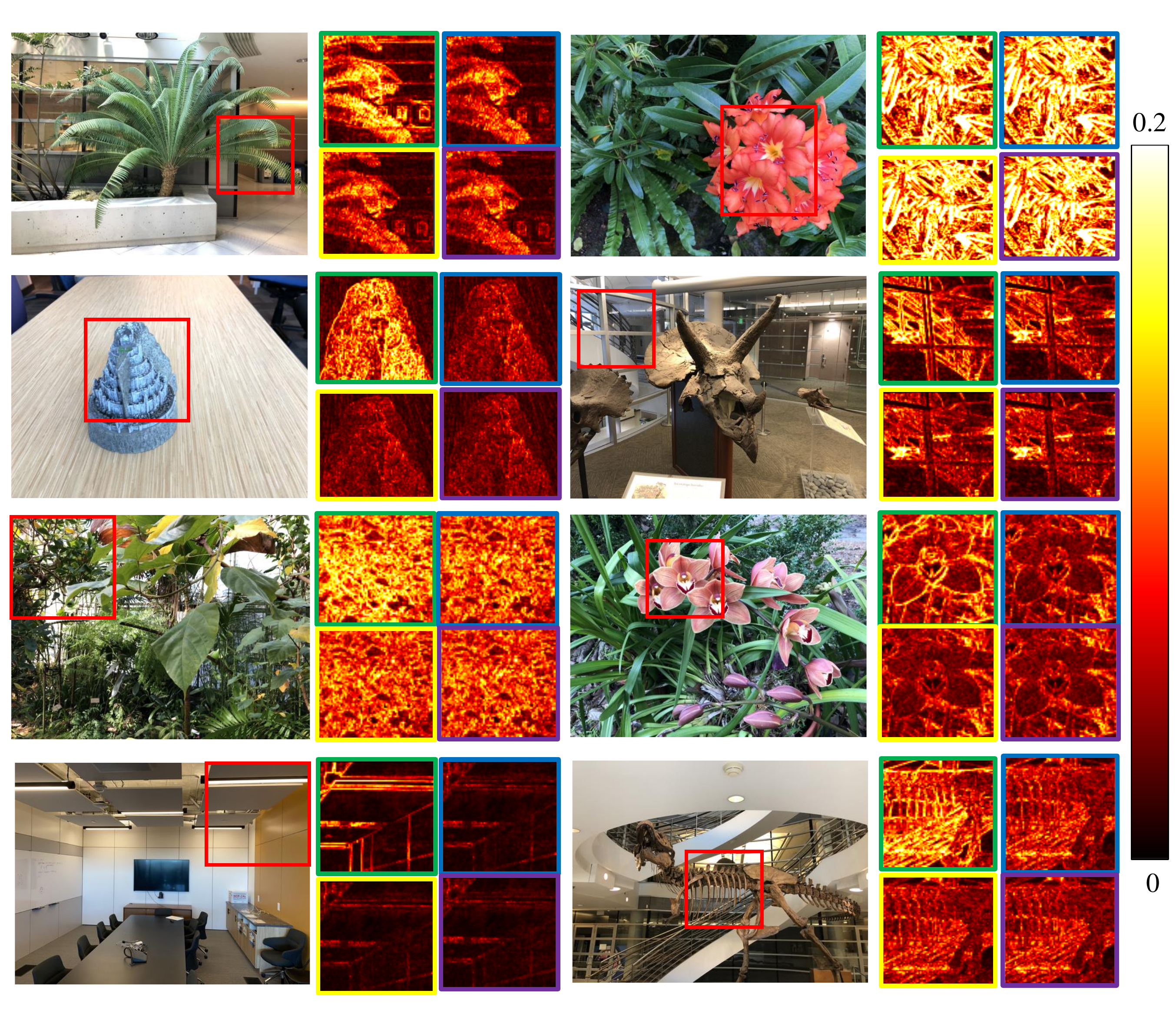}
    \caption{Visualization of rendered images for each phase shown in Table~\ref{tab:ablation}. The green, blue, yellow, and purple box are the error map of NeRF, NeRF + IE, NeRF + IE + OD, and NeRF + IE + OD + PRD, respectively.  }
    \label{fig:my_label}
\end{figure*}

%% file: supplementary/qualitative.tex
\section{Qualitative Results}

We report some qualitative results of experiments in the main paper. Figure~\ref{fig:npp_tanks} compares NeRF++ and our model in a tanks and temples~\cite{knapitsch2017tanks} dataset. Figure~\ref{fig:fisheye_render} compares NeRF++ and our model in two fish-eye scenes. 

\begin{figure*}
    \centering
    \resizebox{0.95\textwidth}{!}{
    \includegraphics{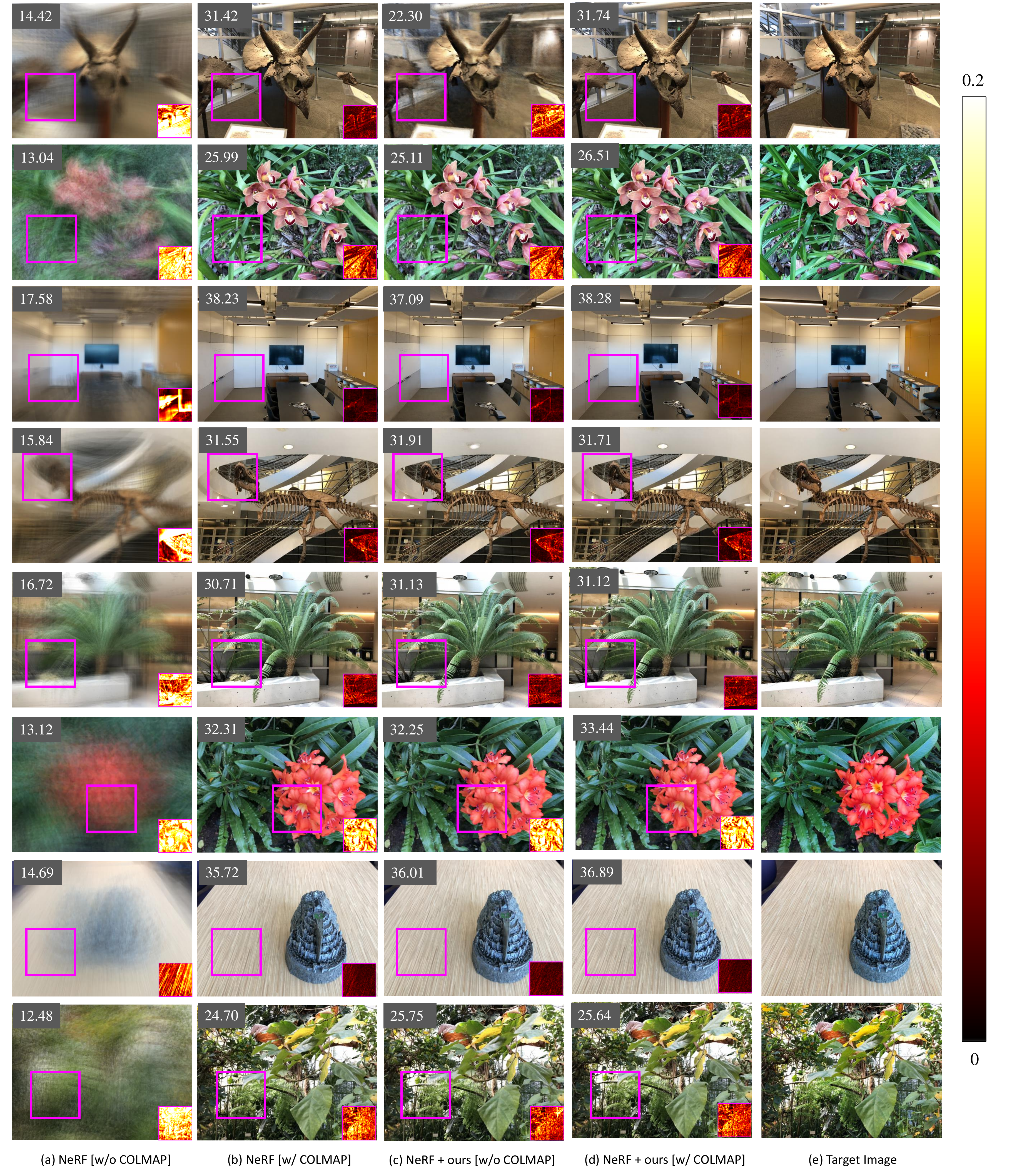}
    }
    \caption{The red block visualizes rendered images without calibrated camera information. Although our model is trained without camera information, our model shows comparable performance with NeRF, trained with COLMAP camera information. The blue block visualizes the rendering of NeRF using COLMAP camera information. For each subfigure, PSNR is shown on the upper left. }
    \label{fig:wo_colmap}
\end{figure*}

\begin{figure*}
    \centering
    \resizebox{0.8\textwidth}{!}{
    \includegraphics{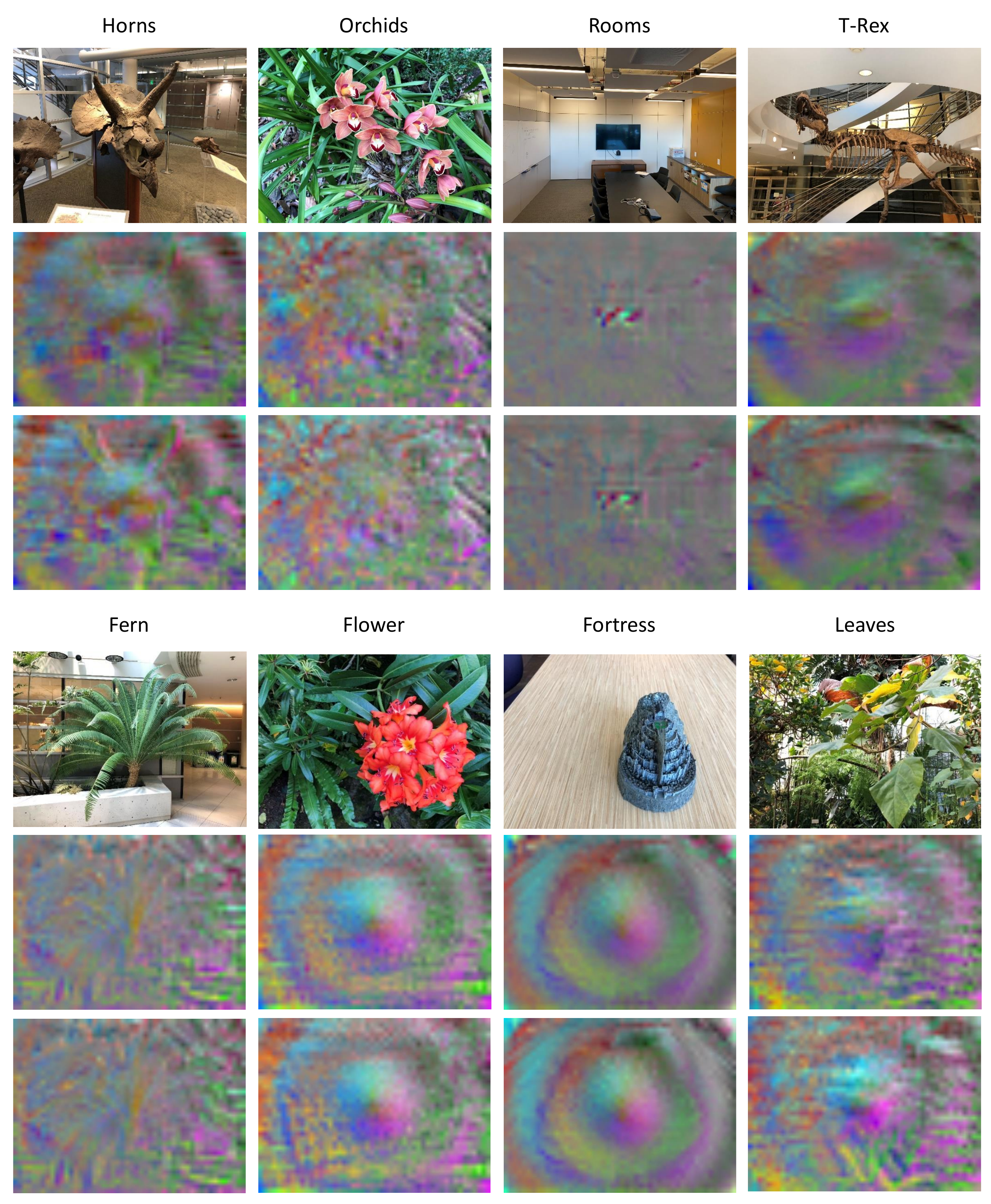}
    }
    \caption{Visualization of captured non-linear distortions of our trained camera model. We have observed circular patterns in all the scenes. The second row and the forth row are captured ray offset distortions, and the third row and the fifth row are captured ray direction distortions.}
    \label{fig:distortion}
\end{figure*}

Figure~\ref{fig:wo_colmap} visualizes rendered images of our model when no calibrated camera information is provided. Lastly, Figure~\ref{fig:distortion} visualizes the captured non-linear distortion in for all the scenes in LLFF~\cite{mildenhall2019local} dataset.

